\documentclass[letterpaper, 10 pt, conference]{ieeeconf}  

\IEEEoverridecommandlockouts                              

\usepackage{graphics} 
\usepackage{epsfig} 
\usepackage{mathptmx} 
\usepackage{times} 
\usepackage{amsmath} 
\usepackage{amssymb}  
\usepackage{cite} 

\title{\LARGE \bf
Ankle Torque During Mid-Stance Does Not Lower Energy Requirements of Steady Gaits
}

\author{Mike Hector, Kevin Green, Burak Sencer, Jonathan Hurst
\thanks{This work was supported by DARPA contract W911NF-16-1-0002 and NSF Grant No. 1314109-DGE. \newline
\- All authors are with the School of Mechanical, Industrial, \& Manufacturing Engineering at Oregon State University, Corvallis, OR, USA. Email
        {\tt\small \{hectorm, greenkev, burak.sencer, jonathan.hurst\}@oregonstate.edu }}%
}

\begin{document}

\maketitle
\thispagestyle{empty}
\pagestyle{empty}

\begin{abstract}

In this paper, we investigate whether applying ankle torques during mid-stance can be a more effective way to reduce energetic cost of locomotion than actuating leg length alone. 
Ankles are useful in human gaits for many reasons including static balancing.
In this work, we specifically avoid the heel-strike and toe-off benefits to investigate whether the progression of the center of pressure from heel-to-toe during mid-stance, or some other approach, is beneficial in and of itself. 
We use an “Ankle Actuated Spring Loaded Inverted Pendulum” model to simulate the shifting center of pressure dynamics, and trajectory optimization is applied to find limit cycles that minimize cost of transport. 
The results show that, for the vast majority of gaits, ankle torques do not affect cost of transport. 
Ankles reduce the cost of transport during a narrow band of gaits at the transition from grounded running to aerial running. 
This suggests that applying ankle torque during mid-stance of a steady gait is not a directly beneficial strategy, but is most likely a path between beneficial heel-strikes and toe-offs.
\end{abstract}

\section{Introduction}

While ankles are not strictly necessary for bipedal locomotion, they are ubiquitous in both bipedal animals and robots \cite{ATRIAS}.
Ankles enable a low effort method of stabilized standing by controlling the position of the center of pressure.
Roboticists designing bipedal robots take advantage of this to allow their robots to stand still; however, it is unclear how ankles should be used during locomotion \cite{alexander_animals}.
Work in biomechanics has shown that ankles have an important role in locomotion including impacting the ground with the heel to soften ground impacts and pushing off at the toe to accelerate the leg into swing.
This is seen in humans along with a transition of the center of pressure from heel to toe during mid-stance.
In addition to being a path between beneficial heel-strikes and toe-offs, the progression of the center of pressure also results in the application of mid-stance ankle torque, which affects the dynamics of walking and running \cite{Holowka2018}.
However, it is unknown whether this progression is beneficial to the gait itself or if it is an emergent feature of heel-strike and toe-off.

This paper presents a first principles approach to understanding mid-stance ankle function.
Mid-stance ankle actuation, not toe-off or heel-strike, is examined to understand how the forces of the leg and ankle can work together to reduce the cost of locomotion.
The leg model is as simple as possible, while retaining the hybrid dynamic and nonlinear nature of legged locomotion.
An actuated spring loaded leg model, commonly used to model leg dynamics, is extended to include an actuated ankle and foot. 
Torque applied at the ankle joint gives control authority to shift the center of pressure only between heel and toe to ensure that the foot stays flat on the ground. 
Through optimization on the model, we arrive at energy efficient trajectories that produce steady gaits for different speeds, body heights, and model parameters.
We developed a measure of ankle utility to describe the energetic benefit of using the ankle. 
We show that mid-stance ankle actuation is often not significantly useful within the constraints of our reduced-order model and hypothesize other reasons for ankle utility in animals and robots.

\begin{figure}[t]
  \centering
  \includegraphics[width = 0.9\columnwidth]{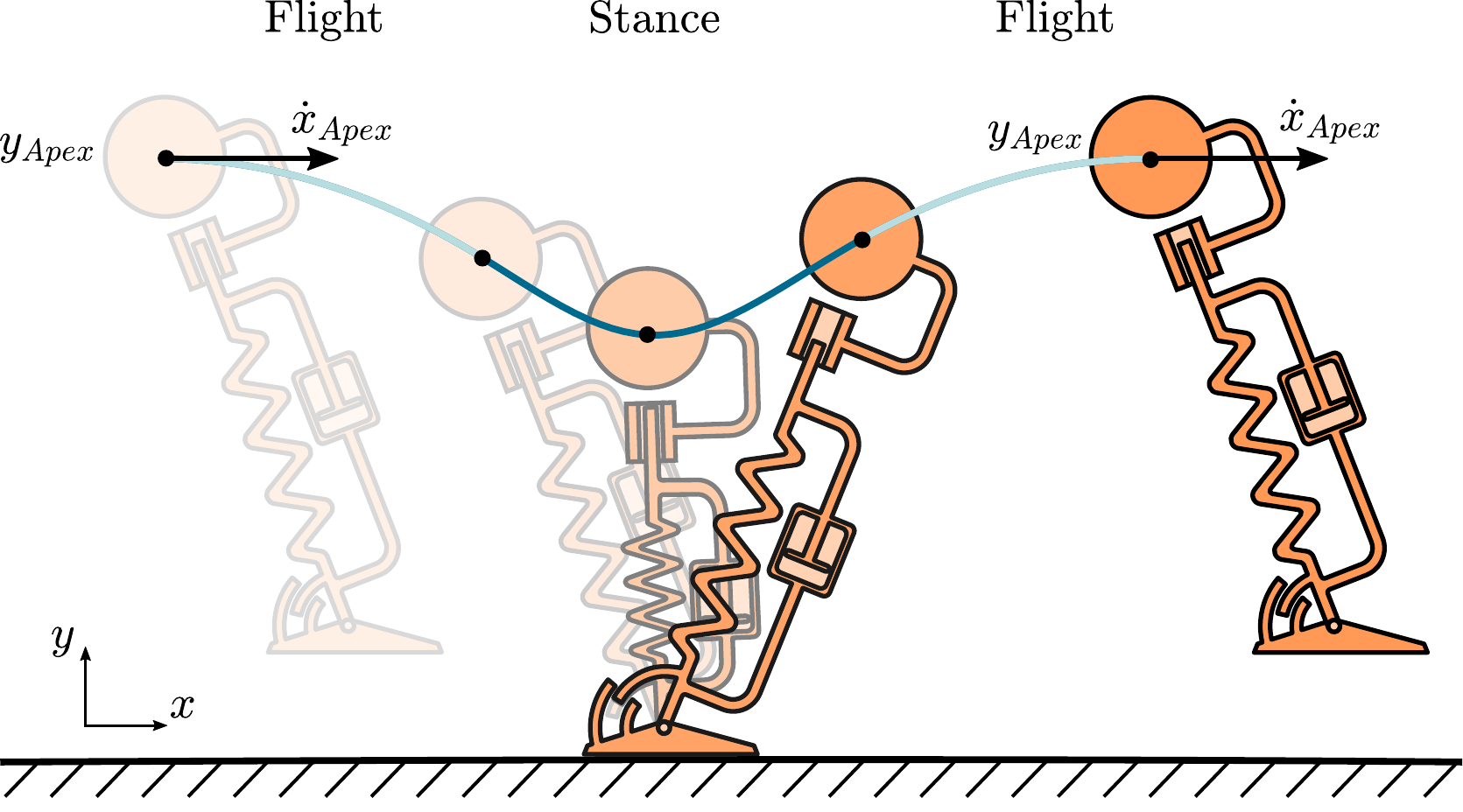}
  \caption{A single optimal cycle in a periodic gait of the Ankle Actuated Spring Loaded Inverted Pendulum.}
\label{fig:aaslipCycle}
\end{figure}

\section{Previous Work on Ankles and Locomotion}

Nature provides examples of legs and ankles with vastly different morphologies; but in general ankles are responsible for applying forces on the body during stance.
We define the ankle as the most distal active joint which enables the center of pressure to be shifted within the support polygon under the foot.
Two common anatomical classifications of leg morphology, plantigrades and digitigrades, have this definition of ankles.
Plantigrades, such as humans, walk with their feet flat on the ground through most of stance and shift their COP within the foot.
In contrast, digitigrades, such as ostriches, walk on their toes, or digits, with raised feet throughout most of stance and shift their COP using their toe \cite{alexander_animals}.
While these ankles seem very different, they can both shift the center of pressure to provide a force on the body during stance \cite{Holowka2018}.

Ankles in humans and animals are understood to make walking and running more efficient primarily by softening ground impacts, propelling the leg forward at lift-off, extending the stride length, and maintaining balance while standing \cite{alexander_animals}.

    Heel-strikes in humans reduce ground impact forces by using soft feet while tendons store energy.
    As the foot comes in contact with the ground, an impact occurs. About 20\% of the collision energy is dissipated through the soft footpad in the human heel \cite{Gefen2001}.
    Additionally, elastic energy is stored in the Achilles tendon \cite{AlexanderCoT}.
    The elastic energy is released during toe-off, which helps accelerate the leg into swing. 
    
    The push-off of the foot corresponds to the largest activation of the ankle during stance, as energy is released in a `catapult' that launches the foot \cite{Lipfert2014}.
    That energy is then transferred up the leg, through the knee and hip, and to the trunk, resulting in forward motion of the body \cite{Meinders1998a}.
    Both heel-strike and toe-off are beneficial to walking gaits, but to do both in a single stride, the center of pressure must be shifted during mid-stance.

    Shifting the center of pressure (COP) is beneficial in humans because it extends the length of a stride which lowers the cost of transport.
    For humans, shifting the COP can be thought of as continuously changing leg placement.
    Landing at the heel and leaving at the toe effectively lengthens the stride \cite{Lieberman2015}.
    To this advantage, the roll-over model was developed, which imagines an arc of a wheel at the bottom of the leg \cite{kuo_OG_rollover}. 
    This rolling progression of the COP from heel to toe shows a decrease in metabolic cost of locomotion in humans \cite{Adamczyk2013}.
    While this can explain the benefits of shifting the COP, it does not give good intuition for the effect of the ankle actuation on the body or the interaction between leg and ankle dynamics.

The first principles approach to modeling legged locomotion commonly uses either the Inverted Pendulum (IP) or Spring Loaded Inverted Pendulum (SLIP) reduced order model to implement stance leg dynamics and to study energy cycles  \cite{Raibert_theGoodBook}.
    The IP and SLIP models encode nonlinear hybrid passive dynamics with few parameters and simple leg morphologies.
        SLIP models can demonstrate steady state locomotion through equilibrium gaits, which can be produced by the correct choice of touchdown angle \cite{Hubicki2015b}.
        In legged locomotion, when a gait has the same height and velocity at the start and end of a cycle (Fig. \ref{fig:aaslipCycle}), a limit cycle called an equilibrium gait has been achieved  \cite{Srinivasan2011FifteenOO}.
        To apply these models to legged robots, damping and actuation can be added so that leg actuation strategies can be investigated.
        Many variants of these models have been created to understand the dynamics of locomotion, some of which also include models of feet and ankles \cite{sharbafi_bigBook}.
        Some leg models have been extended to include ankles by modeling them as rolling wheels and passive torsion springs \cite{Adamczyk2013, Ahn2006}.
        However, there has been no previous work examining how an actuated ankle contributes to energetically efficient locomotion in the context of one of these reduced order models.
        
    This paper investigates how the ankle torque and resulting force on the body affects the overall leg and ankle trajectories to improve the energy efficiency of locomotion.
        We seek to understand if the resulting force during stance contributes to energy efficiency or if heel to toe COP progression is simply a path between beneficial heel strikes and toe offs with net stride length benefits.

\section{Modeling Mid-stance Dynamics of Ankles}

We aim to understand the mid-stance dynamics of legs and ankles by combining the actuated spring loaded inverted pendulum leg model with an ankle actuator.
This model, shown in Fig. \ref{fig:modelFig}, combines features of actuated SLIP models with a simple implementation of mid-stance ankle actuation to gain insight into how shifting COP can improve energy efficiency during stance.

\subsection{SLIP and Actuated SLIP models}
    To understand the trade-offs between leg and ankle actuation, we use a leg model that includes an actuator in series with a damped leg spring.
    The actuated SLIP (ASLIP) dissipates energy through damped spring compression and actuates leg extension through the spring set point. 
    To reduce parameters and complexity of the model, we model the leg actuator by bounding the acceleration of leg extension \cite{Hubicki2015b}.
    These limits create a sense of actuator dynamics without modeling an explicit motor inertia or gear ratio.
    This produces leg dynamics of the form
    \begin{equation}
    \label{ForceLeg}
    F_{leg} = k(r_0 - r) + c(\dot{r}_0 - \dot{r}) 
    \end{equation}
    where $k$, $c$, $r$, $r_0$ are the stiffness, damping, length of the leg, and leg spring set point, respectively.

\subsection{Actuated SLIP with Ankle}
\begin{figure}[t]
  \centering
  \vspace{6 pt}
  \includegraphics[width = 0.9\columnwidth]{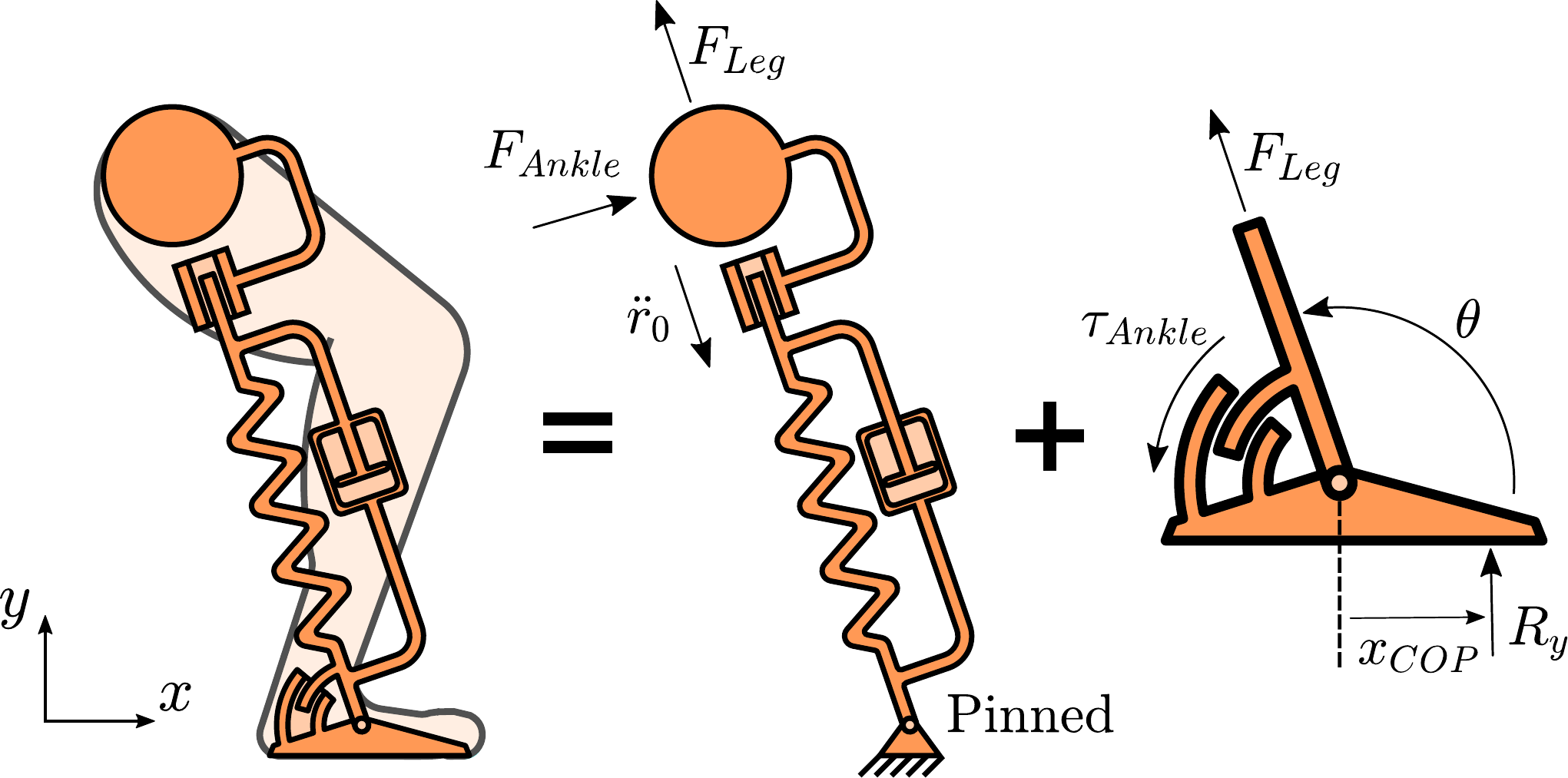}
  \caption{Combined dynamics of the leg and ankle are implemented as a spring mass model with an actuated ankle. The torque at the ankle creates a constrained force on the center of mass that can help propel the model forward through stance.}
\label{fig:modelFig}
\end{figure}

    An ankle is added so that torque at the joint will change the COP and cause a net force on the body (Fig. \ref{fig:modelFig}).
    This combined model is called the Ankle Actuated SLIP or AASLIP.
    The ankle torque, $\tau_{ankle}$, shifts the COP between the heel and toe, but it is strictly limited from applying more torque.
    This keeps the foot flat on the ground and models the force produced by a shifting COP.
    The ankle torque is limited by the COP, which is dependent on leg force, $F_{leg}$, length of the foot, $l_f$, and the leg angle, $\theta$, by
\begin{equation}
    x_{COP} = \frac{-\tau_{Ankle}}{F_{Leg} \sin \theta} 
    \label{eq:COP}
\end{equation}
with limits
\begin{equation}
    -\frac{l_f}{2} \leq  x_{COP} \leq \frac{l_f}{2} .
    \label{eq:COPbound}
\end{equation}
    No ankle torque results in the COP remaining at the ankle axis; thus, the AASLIP model collapses to the ASLIP model when $\tau_{Ankle} = 0$.
    This constrained torque at the ankle can be translated to an ASLIP pinned to the ground with the ability to apply a constrained force on COM perpendicular to leg length direction with magnitude $F_{Ankle} = -\tau_{Ankle}/r$ as shown in Fig \ref{fig:modelFig}.
    This produces the full dynamics equations,
    \begin{align}
        \ddot{x} &= \frac{x F_{Leg}}{mr} - \frac{y F_{Ankle}}{mr} \\
        \ddot{y} &= \frac{y F_{Leg}}{mr} + \frac{x F_{Ankle}}{mr} - g
    \end{align}
    where $m$ is the body mass and $g$ is the gravitational acceleration.
    Because this model is pinned to the ground, friction cone constraints are not included.
    Generally, trajectories generated in this study do not have shallow leg angles that would cause feet to slip during stance.
    
\subsection{Nominal Model Parameters}
Model parameters are chosen to reflect human characteristics in a non-dimensionalized parameter space.
The parameters ($m=1$, $g=1$, $k=20$, $c=0.4$, $l_0 = 1$, $l_f = 0.15$) are non-dimensionalized by the Froude number to make them dimensionless and applicable to many walking systems \cite{Hubicki2015b}.
Stiffness, $k$, and damping, $c$ are chosen to reflect values found in human locomotion studies, and $l_0$ is the nominal length of the leg \cite{Lipfert2014}.

\subsection{Model Strengths and Limitations}
The AASLIP model captures some hypothesized features of leg-ankle mid-stance dynamics while ignoring heel-strike, toe-off, and energy storage.
    For instance, the ankle can apply torque to extend the distance traveled during stance. 
    By using the ankle to push the body through stance, a larger leg angle, $\theta$, at touchdown can be chosen resulting in farther distance travelled during stance. 
    However, this is different from the stride length increase in humans, which is due to changing contact from heel to toe.
    In the AASLIP, the stride length change is caused by the force on the body produced by the ankle.
    
    The AASLIP is a representative model of midstance leg-ankle actuation in both plantitgrades and digitigrades. 
    Since it is focused on the net force from the ankle on the body with the foot firmly on the ground, rather than modeling the extension of the ankle, it can be used to analyze the mid-stance dynamics of any leg-ankle system in stance.
    
    Using the AASLIP model allows us to analyze the interplay of leg and ankle forces on the body.
    Additionally, because it is an actuated locomotion model, it can be used to analyze the energetics of steady gaits to give insights into how legs and ankles work together to make locomotion more efficient.


\section{Calculating Mid-stance Ankle Utility using Optimization}

We calculate the energetic benefit of using ankles during mid-stance by using trajectory optimization to find and compare the optimal equilibrium gaits of both the model with the ankle (AASLIP) and the model without an ankle (ASLIP).
Energy efficiency is used because there is evidence that animals minimize power in steady gaits to increase energy economy over long distances \cite{AlexanderEmin}.
The energy required to complete an equilibrium gait is commonly non-dimensionalized to a quantity which is applicable to all kinds of locomotion called the Cost of Transport ($CoT$), which is defined as 
\begin{equation}
CoT = \frac{E_{req}}{mgd}
\end{equation}
where $E_{req}$ is the energy input required for the gait, $m$ is the body mass, $g$ is the gravitational acceleration, and $d$ is the distance traveled in one cycle \cite{AlexanderCoT}.
Ankle benefit is determined by comparing equilibrium gait CoTs of the AASLIP and ASLIP with matching model and gait parameters. 
This new quantity called ankle utility, $U$, is defined as the percent decrease in $CoT$ to complete the cycle when the ankle is present or
\begin{equation}
U = 100 \cdot \frac{CoT_{No \ Ankle} - CoT_{Ankle}}{CoT_{No \ Ankle}} .
\end{equation}
An ankle utility, $U$, greater than 0 means that the ankle is energetically beneficial.
Since no ankle actuation collapses the AASLIP model to the ASLIP model, the ankle utility must always be greater than 0.
To calculate ankle utility, a trajectory optimization is posed to determine minimal $CoT$ trajectories which produce equilibrium gaits.

\subsection{Objective Function}
The optimization objective is to minimize $CoT$ using a cost function which describes both positive actuator work and thermal losses.
This is implemented as
\begin{equation}
    \begin{aligned}
E_{req} &= \int_{0}^{T} max(0,\ (1-\alpha) P_{Leg} + \alpha R_{Leg} F_{Leg}^2 ) \ dt \\
&+ \int_{0}^{T} max(0,\ (1-\alpha) P_{Ankle} + \alpha R_{Ankle} \tau_{Ankle}^2 ) \ dt    \end{aligned} 
\end{equation}
where
\begin{equation}
P_{Leg} = F_{Leg} \cdot \dot{r}_0
\end{equation}
\begin{equation}
P_{Ankle} = \frac{\tau_{Ankle} (\dot{x} y - x \dot{y})}{r^2} .
\end{equation}
Positive mechanical actuator work is modeled because negative work is difficult to recover \cite{Remy2015}.
Integrating the positive instantaneous mechanical power of the actuator, $P_{i}$, over time gives the positive work done by the actuator.
A smoothing function for $max(0, \ x)$ of the form 
\begin{equation}
    max(0, \ x) = \frac{x + \sqrt{x^2 + \epsilon^2}}{2}   
\end{equation}
where $\epsilon = 10^{-6}$ is used to eliminate negative contributions of mechanical power and maintain a differentiable objective function.

Thermal losses or electrical losses are used to penalize high  force  in  the  actuator by using a model based on DC motors.
In a DC motor thermal loss is the power that is dissipated in the resistance of the motor windings.
Since resistive power losses are proportional to the square of current and current is proportional to the actuator force or torque output, the thermal losses are proportional to the square of actuator effort. 
We use $R_{Leg}$ and $R_{Ankle}$ to represent the thermal loss coefficient in the leg and ankle, respectively.

When representing a system with both leg and ankle motors there exists a balance between each actuator's electrical losses which is influenced by motor selection and gear ratio \cite{YevOptimalControl}. 
The selection of actuator and transmission is influenced by many factors beyond the thermal losses in the system, such as peak torque, reflected inertia, maximum velocity, and maximum acceleration.
We choose this balance to match the bipedal robot Cassie, designed by the Oregon State University Dynamic Robotics Laboratory.
This is a conservative selection because the ankle motor has a much higher gear ratio compared to the leg extension motor, which reduces the thermal losses of the ankle.
The thermal loss coefficients are determined by non-dimensionalizing a function of Cassie's actuator forces and power losses into the reduced order model space. 
This is done for both the leg and ankle motors to calculate $R_{Leg} = 0.028 \ [m^{-1}l_0^{3/2}g^{3/2}]$ and $R_{Ankle} = 0.50 \ [m^{-1}l_0^{-1/2}g^{-1/2}]$. 
 Because $R_{Leg}$ translates force to power loss, and $R_{Ankle}$ translates torque to power loss, they cannot be directly compared.

In addition to setting the magnitude of the actuator electrical losses relative to each other, the balance of total mechanical losses to total electrical losses is also important because it describes many configurations of actuators and gearboxes at different motor scales.
To explore this balance, we introduce a convex combination parameter, $\alpha$, with the property that varying $\alpha$ from 0 to 1 represents the cost function changing from only mechanical power to only thermal losses.
Only the one parameter is needed because the total magnitude of the objective function has no effect on the optimal solution.
Additionally, when $\alpha = 0.5$ the mechanical power and thermal loss terms are balanced to match the nominal case with parameters that match Cassie.

\subsection{Constraints}
Constraints are imposed on this problem to ensure feasible dynamics, ankle torques, leg accelerations, and equilibrium gaits.
Dynamics are implemented as constraints through trapezoidal direct collocation with 30 knot points \cite{kellyTrajOpt}.
To avoid modeling a hybrid system, the flight phases are formed as constraints on the initial and final conditions of stance. These constraints guarantee ballistic trajectories into and out of stance such that apex conditions are reached.

Periodic equilibrium gait constraints are created by matching the states at the initial and final apexes of the trajectories where there is no vertical velocity, thus creating a single stride that could be repeated to create a stable gait.

During the trajectory, ankle torques are constrained by Eq. \ref{eq:COP} and Eq. \ref{eq:COPbound} so that the COP stays between the heel and toe.
Similarly, the acceleration of the leg motor, length of the leg, and force in the leg are bounded between $-1 \leq \ddot{r}_0 \leq 1$, $.5 \leq r \leq 1$, and $0 \leq F_{Leg} \leq 5$, which respects values found in other locomotion studies \cite{Hubicki2015b}.

\subsection{Implementation}
The nonlinear optimization problem posed can be solved efficiently using a standard nonlinear program solver.
Matlab Symbolic Math Toolbox was used to generate analytical gradients of the objective and constraints.
These are sent to fmincon, a nonlinear constrained optimization program, and solved using a sequential quadratic programming algorithm.
Resulting trajectories are consistently found in less than 10 seconds on a desktop computer and are within a constraint tolerance of $10^{-9}$.
With this framework, $CoT$ can be calculated for both the ankle and no ankle models. The ankle utility was then recorded as the percent decrease of the $CoT$ for the AASLIP over the ASLIP.

\section{Changing Parameters has minimal effect on Ankle Utility}
With the goal of finding scenarios in which ankles are useful, ankle utility is calculated while varying gait apex height, gait apex velocity, $\alpha$, and leg damping. 
While there is little ankle utility in the nominal case, when varying apex height, there is a narrow band of apex heights with significant ankle utility. 
We then vary other parameters along with apex height to show that ankle utility is only significant in a narrow band of apex heights which corresponds to the transition from grounded running gaits with minimal aerial phase to running gaits with a prominent aerial phase.

\subsection{Sample Optimal Trajectories}
\begin{figure}[thpb]
  \centering
  \includegraphics[scale=.4]{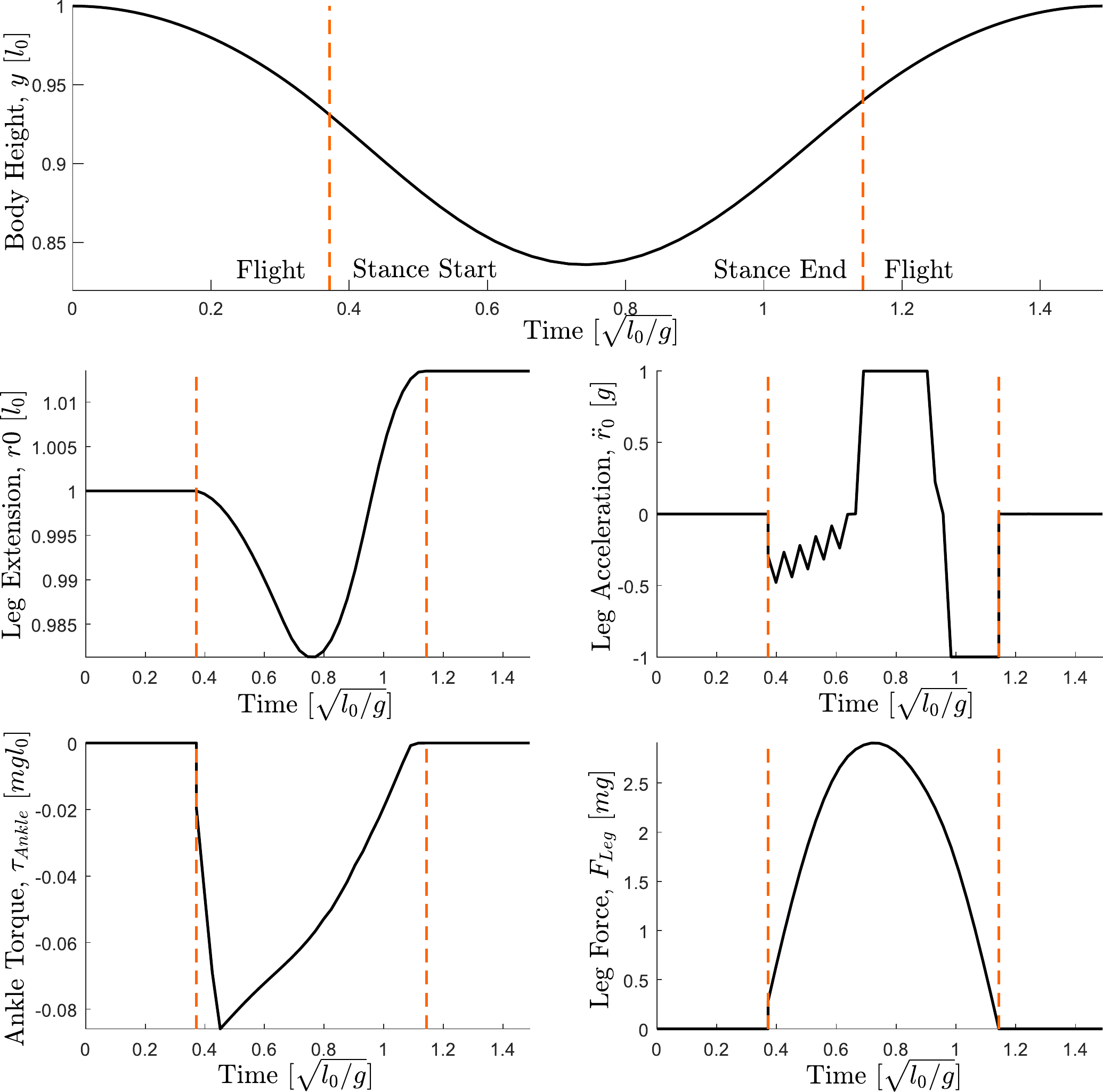}
  \caption{Sample trajectories from the ankle actuated SLIP with nominal parameters. The dip in leg extension is the result of balancing positive mechanical work and thermal losses. The negative $\tau_{ankle}$ helps to propel the body in the direction of motion, but its magnitude is very small relative to ground reaction forces.}
\label{fig:traj}
\end{figure}

Fig. \ref{fig:traj} shows a sample trajectory for an AASLIP model with nominal model parameters ($\alpha = 0.5$). 
The leg dynamics demonstrate an expected balance of positive work and force in the leg, while the ankle shows little activation.
A trade-off between thermal losses and mechanical work is seen in the $r_0$ trajectory dip, which minimizes just before mid-stance when the leg force is highest.
After the middle of stance, a bang-bang response on $\ddot{r}_0$ adds energy through leg extension to power lift off.
Throughout the stride, the ankle pushes on the body in the direction of motion; however, the magnitude of the force is very small compared to the force in the leg.

$CoT$ from this trajectory will be combined with $CoT$ from an optimization on the ASLIP model to find whether the force from the ankle can improve energy efficiency.

\subsection{Effects of Apex Height on Ankle Utility}
\begin{figure}[thpb]
  \centering

  \includegraphics[scale=.4]{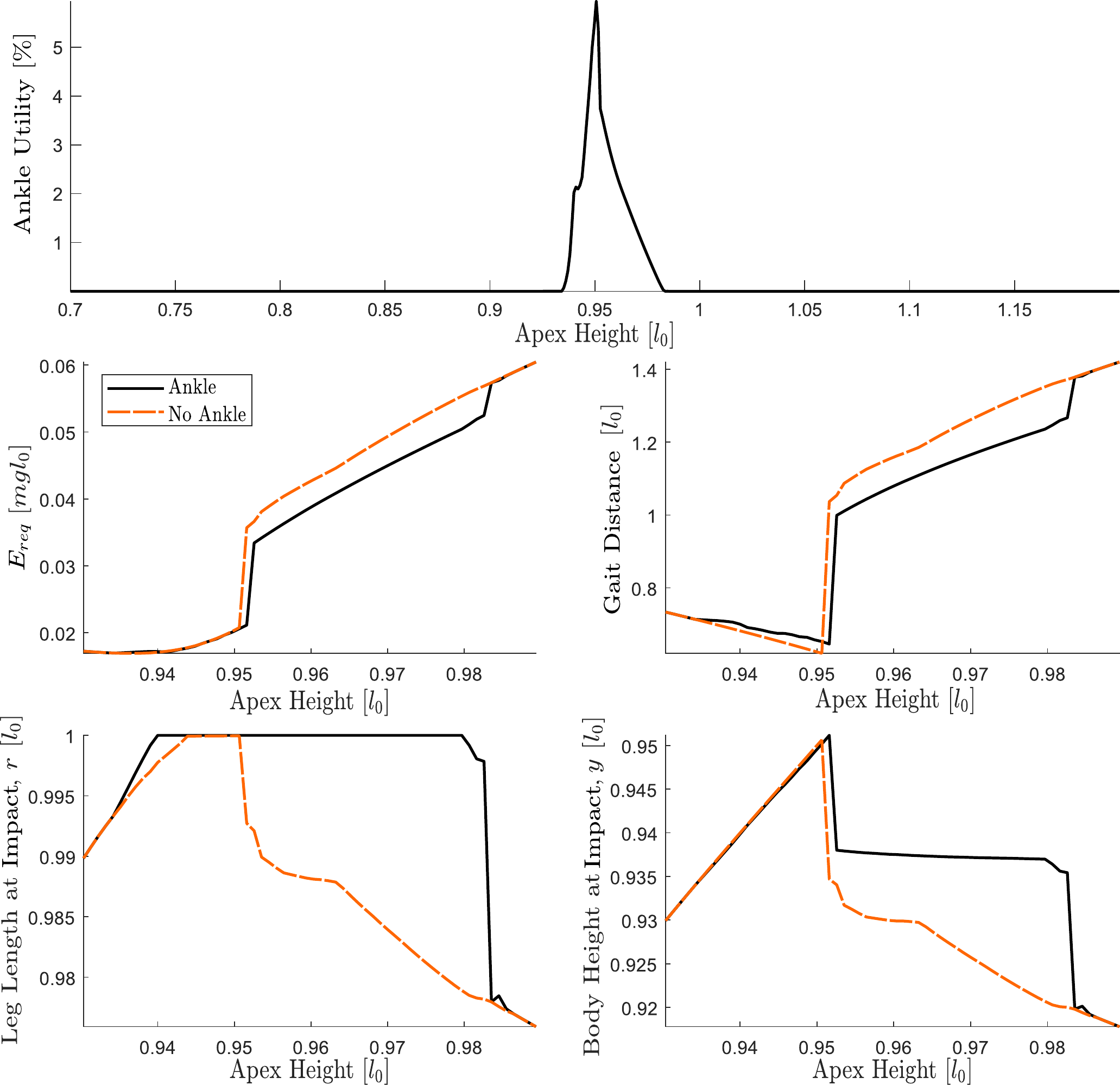}
  \caption{Ankle utility is significant in only a narrow band of apex heights and minimal elsewhere. 
  In this range, actuating the ankle allows a more extended leg length at impact with the ground, which reduces energy lost in the leg damper.
  Below an apex height of $l_0 = 0.95$, this results in a farther gait distance with similar energy requirements; whereas, above that apex height, the body height at impact is increased which decreases the energy required.}
\label{fig:ah}
\end{figure}

Ankle utility is calculated for a range of feasible apex heights to show that it is only significant 
during gaits with apex heights near $0.95l_0$.
To aid a clear understanding of why ankle utility is high in this band, we consider the case when only mechanical work and no thermal losses ($\alpha = 0$) are included.
Fig. \ref{fig:ah} shows the parameter sweep results.

An apex height range between $0.6 l_0$ and $1.2 l_0$ is chosen to produce feasible trajectories with respect to the bound on force in the leg.
Apex heights less than $0.95 l_0$ produce gaits with minimal flight phase. While this might be problematic for a monopod hopper, which would not be able to move its leg to the new touchdown position in time, adding a second leg would allow for a feasible grounded run in which only one leg is on the ground at any time.
The result of this parameter sweep shows that there is a small band of significant ($>5 \%$) ankle utility, with negligible ankle utility outside that band.
\begin{figure*}[t]
  \centering
  \vspace{6 pt}
  \includegraphics[width=\textwidth]{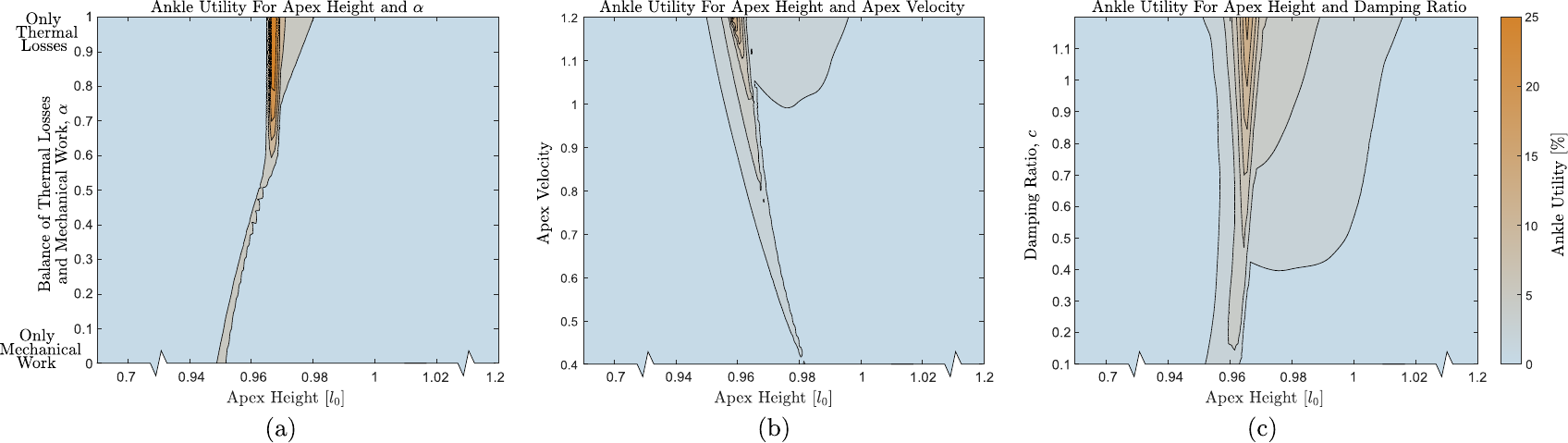}
  \caption{Plots of ankle utility for ranges of two parameters: (a) apex height and alpha, (b) apex height and apex velocity, (c) apex height and damping. These contours show that ankle utility is significant only in narrow bands close to $l_0=0.95$, which indicates that apex height is the dominant parameter for ankle utility. Thus there is little energetic benefit to ankle actuation except during transition from grounded running gaits to running gaits.}
\label{fig:contour}
\end{figure*}

Ankle intervention in this band corresponds to either lowered energy input or longer stride distance.
At all points in this band, ankle intervention increases touchdown leg length which reduces losses in the damper.
This happens because the ankle can add energy to the system without direct losses; whereas, the leg must extend through the damper which dissipates energy. 
With less leg extension needed, the SLIP can increase leg length at touchdown, which corresponds to a shorter flight phase.

When apex height is below the apex height of maximum ankle utility, the high ankle utility is caused by an increase in stride length with only small increases in energy requirements.
The ankle actuation allows a more shallow leg angle to be chosen while maintaining a similar touchdown height and leg length. 
This means that more distance is traveled throughout the stride with only a small increase in energy required.

Above the apex height of maximum ankle utility, the previous strategy breaks down in favor of using the ankle actuation to avoid leg thrust so that the SLIP can touchdown with its body closer to apex height and dissipate less energy in the damper. This decreases the energy required and the distance traveled, but results in a net decrease of the $CoT$ compared to no ankle.

Apex heights at the beginning of this band of high ankle utility ($>5 \%$) and below have minimal or even no flight phase, which makes them analogous to grounded running gaits.
Thus, this band most closely represents the transition from grounded gaits to aerial running gaits. 
Outside this band, most notably in high apex height running gaits, it is clear that the ankle makes little contribution to the energetics of the system. 
With this knowledge, we explore other parameters alongside apex height to see how ankle utility is affected.

\subsection{Apex Height is the Dominant Parameter for Increasing Ankle Utility}

To show that apex height is the dominant parameter for increasing ankle utility, we present two dimensional grids of parameter values and their resulting ankle utilities. 
In Fig. \ref{fig:contour} apex height is varied along with $\alpha$, damping, and apex velocity to create contours of ankle utility at different parameter values. 
The results indicate that mid-stance ankle actuation is not energetically beneficial except in a narrow band of apex heights.
%

Fig. \ref{fig:contour} (a) shows ankle utility for values of apex height and $\alpha$. Varying $\alpha$ from 0 to 1 describes a range of systems with power consumptions that are a spectrum between purely mechanical work and purely thermal losses. 
When $\alpha = 0.5$ the nominal balance of mechanical work and thermal losses reflects the hardware used in Cassie.
As with the nominal $\alpha =0.5$ case in Fig \ref{fig:ah}, we see a trend of high utility in a narrow band around $0.95l_0$.
There is some utility outside that band, but it is less than 3\%.
Generally, as $\alpha$ increases, ankle utility increases because the ankle helps decrease the force in the leg through stance by allowing a higher touchdown height.
Since increasing $\alpha$ puts a heavier weight on high leg forces, it becomes more advantageous to use the ankle.

This trend continues in Figs. \ref{fig:contour} (b) and (c). 
Neither contour shows significant ankle utility outside of the transition from grounded running to aerial running gaits. 
In the case of damping, the narrow range is widened at higher values as the model avoids thrusting the leg, which dissipates energy in the damper, in favor of using the ankle to power through stance. 
For Fig. \ref{fig:contour} (b) an increase in apex velocity means hitting the ground at a higher velocity which means more energy will be dissipated in the damper. 
Again, adding energy by actuating the ankle avoids leg thrust, making it an advantageous strategy. Additionally, very little ankle utility is found outside of the band around apex height of $0.95l_0$ for any set of parameters, and the regions of low ankle utility correspond to low ankle usage (zero ankle torque).

\subsection{Relevance to More Complex Systems}
This model has shown that ankle torque is minimally useful during mid-stance for steady gaits. 
Applying this to more complex systems, such as human walking, suggests that the advantages of heel-strike and toe-off dominate ankle dynamics. 
This implies that the COP shift from heel to toe is most likely a path between beneficial heel-strikes and toe-offs with the added benefit of changing contacts lengthening stride, while the forces on the body caused by the shifting COP do not significantly contribute to locomotion.
For the case of human running, the advantages of energy storage in the Achilles tendon and the net leg extension at toe-off can explain the common mid-foot to toe COP progression, while the ankle's force on the body contributes very little to forward motion.
Additionally, ankle torques could allow a biped to control body pitch during stance without affecting the dynamics of its center of mass.

Because of complications in human morphology and muscle actuation, humans might not be the best match for this model; however, the SLIP model has been a useful basis of designing and controlling legged robots \cite{Apgar2018}. 
In addition to mechanism design, many robots are also actuated by SLIP model inspired controllers. 
From this research, an interesting control strategy emerges for robot ankles: once the foot is on the ground, don’t apply torque until it is time to push off. 
It is better to focus on designing mechanisms with good impact robustness and with enough power to push off to accelerate the leg into swing, rather than designing a mechanism or controller with a specific COP progression in mind.

\section{Conclusions and Future Work}
This paper investigated the role of ankles during mid-stance to improve the energetics of legged locomotion. 
The work was inspired by the COP progressions in humans and animals to study how ankle torque during mid-stance improve the energy economy of locomotion.
To investigate the energetic benefits of these COP progressions, we took a first principles approach to modeling the dynamics of leg-ankle systems.
By extending previously studied models of actuated legs with an ankle, we were able to model the forces on the body produced by ankle torques shifting the COP. 
We applied trajectory optimization to determine a measure of the energetic benefit of using the ankle for a steady gait. 
The result of those optimizations showed that mid-stance ankle torques improve walking and running gaits, but only in a narrow region in the parameter space. 
Thus, in general, mid-stance ankle actuation is not energetically beneficial for walking and running.
However, COP progressions can be useful in humans and animals because of complications such as morphology and established benefits of heel-strike and toe-off.

Future studies of ankles for legged robots could take a similar approach as this study, and by extending the model for heel-strikes or toe-offs, it can show how to use ankles to make walking and running robots more robust and energy efficient.

\addtolength{\textheight}{-12cm}   




\bibliographystyle{IEEEtran.bst}
\bibliography{references.bib}

\end{document}